\documentclass[]{article} 

\usepackage[utf8]{inputenc} 
\usepackage[T1]{fontenc}    
\usepackage{hyperref}       
\usepackage{url}            
\usepackage{booktabs}       
\usepackage{amsfonts}       
\usepackage{nicefrac}       
\usepackage{microtype}      
\usepackage{tikz}
\usetikzlibrary{decorations.pathreplacing} 
\usepackage[numbers]{natbib} 
\usepackage{algorithmic}
\usepackage{algorithm}

\newlength{\myleftmargin}
\usepackage{amsmath}
\usepackage{amssymb}
\usepackage{amsthm}
\usepackage{color}
\DeclareSymbolFontAlphabet{\Bbb}{AMSb}

\newtheorem{theorem}{Theorem}[section]
\newtheorem{lemma}[theorem]{Lemma}

\newtheorem{definition}[theorem]{Definition}

\newenvironment{proofof}[1]{\noindent{\bf Proof of #1:}}{\qed\medskip}

\newlength{\fixboxwidth}
\definecolor{darkgreen}{rgb}{0,0.6,0}

\newcommand{\ca}[1]{{\cal #1}}

\newcommand{\mycdot}{\,\cdot\,}
\newcommand{\Leq}{\,\leq \,}

\newcommand{\R}{\mathbb{R}}
\newcommand{\Rd}{\mathbb{R}^d}

\renewcommand{\a}{\alpha}

\newcommand{\g}{\gamma}

\newcommand{\D}{\Delta}
\newcommand{\e}{\varepsilon}

\newcommand{\lb}{\lambda}

\renewcommand{\t}{\tau}

\renewcommand{\P}{\Phi}

\DeclareMathOperator{\spann}{span}

\newcommand{\eins}{\boldsymbol{1}}

\newcommand{\snorm}[1]{\Vert #1 \Vert}
\newcommand{\mnorm}[1]{\bigl\Vert \, #1 \, \bigr\Vert}

\newcommand{\inorm}[1]{\Vert #1 \Vert_\infty}

\newcommand{\RP}[2]{{{\cal R}_{#1,P}(#2)}}
\newcommand{\RPB}[1]{{{\cal R}_{#1,P}^{*}}}
\newcommand{\RD}[2]{{{\cal R}_{#1,D}(#2)}}

\newcommand{\Rx}[3]{{{\cal R}_{#1,#2}(#3)}}

\newcommand{\clippt} {{{}^{\textstyle \smallfrown}} \hspace*{-1.5ex} t\hspace*{0.5ex}}

\newcommand{\clippfo} {{{}^{^{\textstyle \smallfrown}}} \hspace*{-2.1ex} f}

\newcommand{\fTc}{\clippfo_{D,\lb}}
\newcommand{\fTcn}{\clippfo_{D,\lb_n}}

\newcommand{\fDl}{f_{D,\lb}}

\title{Learning with Hierarchical Gaussian Kernels}

\author{
  Ingo Steinwart \\ 
  Institut für Stochastik und Anwendungen\\
  Universität Stuttgart \\
  Pfaffenwaldring 57, 70569 Stuttgart \\
  \texttt{ingo.steinwart@mathematik.uni-stuttgart.de}\\
  \and
  Philipp Thomann \\
  Institut für Stochastik und Anwendungen\\
  Universität Stuttgart \\
  Pfaffenwaldring 57, 70569 Stuttgart \\
  \texttt{philipp.thomann@mathematik.uni-stuttgart.de}\\
  \and
  Nico Schmid \\
  Bosch Healthcare Solutions \\
   \texttt{Nico.Schmid2@de.bosch.com}
}

\begin{document}

\maketitle

\begin{abstract}
We investigate iterated compositions of weighted sums of Gaussian kernels and provide an
interpretation of the construction that shows some similarities with the architectures of  deep neural networks.
On the theoretical side, we show that these kernels are universal and that SVMs using these 
kernels are universally consistent. We further describe a parameter optimization method 
for the kernel parameters and empirically compare this method to SVMs, random forests, a multiple kernel learning
approach, and to some deep neural networks.
\end{abstract}

\section{Introduction}\label{sec:intro}

Although kernel methods such as support vector machines 
are one of the state-of-the-art methods when it comes to fully 
automated learning, see e.g.~the recent independent comparison
\cite{FeCeBaAm14a}, the recent years have shown that on complex datasets
such as image, speech and video data,
they clearly fall short 
compared to deep neural networks. 

One possible explanation for 
this superior behavior is certainly their deep architecture 
that makes it possible to represent highly complex functions 
with relatively few parameters. In particular, it is possible to 
amplify or suppress certain dimensions or features of the 
input data, or to combine features to new, more abstract features.
Compared to this, standard kernels such as the popular Gaussian 
kernels simply treat every feature equally. In addition, most users of 
kernel machines probably stick to the very few standard kernels, often 
simply because there is in most cases no principled way for 
finding problem specific kernels. In contrast to this, deep neural networks
offer yet another order of freedom to the user by making it easy to 
choose among many different architectures and other design decisions.
This discussion shows that the class of deep neural networks offers
potentially much more functions that may fit well to the problem at hand
than classical kernel methods do. 
Therefore, if the training algorithms are able to find these good hypotheses
while  simultaneously controlling the inherent danger of overfitting (and the user 
picked a good design), 
then the recent success of deep networks does not seem to be so surprising after all.
In particular, this may be an explanation for the types of data mentioned above, for
which an equal and un-preprocessed use of all features may really not be the best idea,
but for which there is also some biological insight suggesting certain building blocks of  neural architectures.
Moreover, the recent success of deep networks indicates that these `ifs' can nowadays much better
be controlled than 20 to 30 years ago.
This naturally raises the question, whether and how certain aspects of deep neural 
networks can be translated into the kernel world without sacrificing the benefits 
of kernel-based learning, namely less `knobs' an unexperienced user
can play with, the danger of getting stuck in poor local minima, a more principled 
statistical understanding, and last but not least, their success in situations in which 
no human expert is in the loop.

Of course, the limitations of using simple single kernels have already been recognized before.
Probably the first attempts in this direction are multiple kernel learning algorithms, see e.g.~\cite{SoRaScSc06a}, which,
in a nutshell,  replace a single kernel by a weighted sum of kernels. The advantage 
of this approach is that finding these weights can again be formulated as a convex objective, while
the disadvantage is the limited gain in expressive power unless the used dictionary 
of kernels is really huge. 
A more recent approach for increasing the expressive power is to construct complex kernels 
from simple ones by composing their feature maps in some form.
Probably the first result in this direction can be found in \cite{ChSa09a}, in which 
the authors described the general setup and considered some particular constructions.
Moreover, this idea was adopted in \cite{StVi13a}, where the authors considered sums of 
kernels in each composition step and established bounds on the Rademacher chaos complexities.
Similarly, \cite{ZhTsHo11a-art} considered such sums in the decomposition step, but they 
mostly restricted their considerations to a single decomposition step, for which they established
a generalization bound based on the pseudo-dimension.
Furthermore, \cite{WiHuSaXi16a-art} 
investigated compositions, in which the initial map is not a kernel feature map, but 
the map induced by a deep network. All these articles also present some experimental
results indicating the benefits of the more expressive kernels. Similarly, \cite{Tang13a}
reports some experiments with a linear SVM as the top layer of a deep network.
Another approach  can be found in 
\cite{mairal:hal-01005489}, where the authors construct 
hierarchical convolutional kernels for image data, and finally 
 \cite{Bach08a} proposed a multiple kernel learning approach, which is also called 
hierarchical kernels. However, besides the name, that paper has  little similarity to our approach.

In this paper we  adopt the   idea of iteratively composing weighted sums of kernels 
in each layer. Unlike the papers mentioned above, however, we focus on sums of Gaussian 
kernels composed with Gaussian kernels. Besides an illustrative interpretation of the construction,
which highlights the similarities to deep architectures, we show that the resulting kernels
are universal and that the corresponding SVMs become universally consistent. 
To the best of our knowledge our paper is thus the first that does not only investigate
some statistical properties of   iterated   kernels, but also considers their approximation properties. 
We further describe
a possible kernel parameter optimization algorithm in detail, and last but not least we report
results from extensive experiments comparing our algorithm
against SVMs, random forests, the hierarchical kernel learning (HKL) of \cite{Bach08a}
and deep neural networks. Here it turns out that 
even for rather small compositions our approach consistently outperforms SVMs and HKL.
Moreover, our experiments indicate an advantage over deep neural networks
for which the architecture is automatically determined by cross validation
and they further show that our approach  performs better than random forests.

%

The rest of this paper is organized as follows: In Section \ref{sec-hier-kernels}
we introduce our class of kernels and in  Section \ref{sec:analysis} we
present their theoretical analysis. Section \ref{sec:param-optimization}
contains the description of the parameter optimization, and the experiments are reported in 
Section \ref{sec:experiments}. The proofs and some auxiliary theorems can be found in the appendix.

\section{Hierarchical Gaussian Kernels}\label{sec-hier-kernels}

%

In this section we introduce the central object of this paper, namely hierarchical Gaussian kernels.
Let us begin by recalling some basic facts on kernels from \cite[Chapter 4]{StCh08}.
To this end, let $k:X\times X\to \R$ 
be a function. Then $k$ is a kernel, if there exists a Hilbert space $H$ and a map 
$\P_0:X\to H_0$ such that 
\begin{displaymath}
   k(x,x') = \langle \P_0(x), \P_0(x')\rangle\, , \qquad \qquad x,x'\in X.
\end{displaymath}
In this case, $\P_0$ and $H_0$ are said to be  a feature map and a feature space of $k$, respectively.
It is well known, that every such kernel possesses a unique reproducing kernel Hilbert space (RKHS)  $H$ 
and this RKHS is a feature space of $k$ with respect to the canonical 
feature map $\P:X\to H$ given by $\P(x) := k(x,\cdot)$, $x\in X$. Moreover, by definition $H$ consists
of functions $h:X\to \R$, and if $k$ is continuous, so are the functions in $H$.
Let us now assume that $X$ is a compact metric space and that $k$ is a continuous kernel. 
Then
 $k$ is said to be universal, if its RKHS $H$ is dense in the space $C(X)$ of continuous functions 
$f:X\to \R$ with respect to the supremums norm $\inorm\cdot$.
Furthermore, we say that $k$  is injective, if its canonical feature map is injective.
Recall that universal kernels are injective, see e.g.~\cite[Lemma 4.55]{StCh08}. 
Moreover, the injectivity is actually shared by all feature maps of $k$, see 
Lemma \ref{injective-kernel-char} for details.
In particular, it is easy to see that 
besides universal kernels many other kernels are injective, too. For example, 
the linear kernel $k(x,x') := \langle x,x'\rangle$ on $\R^d$ is injective since one of its 
feature maps is the identity on $\R^d$, which, of course, is injective. Finally, note that 
if $k$ is injective on $X$ then its restriction $k_{|X'\times X'}$ onto some subset $X'\subset X$
is still injective.


Probably one of the best known universal kernel
 on $X\subset \Rd$ is the standard Gaussian RBF kernel $k_\g$ with width $\g>0$, which is given by
\begin{displaymath}
   k_{\g, X}(x,x') := \exp\bigl( -\g^{-2}\snorm{x-x'}_2^2  \bigr)\, , \qquad \quad x,x'\in X\, ,
\end{displaymath}
where $\snorm\cdot_2$ denotes the standard Euclidean norm on $\Rd$. 
In the following we denote its RKHS by $H_{\g, X}$.
Now, it is well-known 
that this construction
can actually be extended to subsets $Z$ of more general Hilbert spaces $H$, that is, 
\begin{equation}\label{gauss-on-H}
   k_{\g, Z}(z,z') := \exp\bigl( -\g^{-2}\snorm{z-z'}_H^2  \bigr)\, ,  \quad z,z'\in Z  .
\end{equation}
is again a   kernel, whose RKHS will be denoted by $H_{\g, Z}$.
In particular, if we have a map $\P:X\to H$, then 
\begin{equation}\label{hier-prot-kernel}
   k_{\g, X,H}(x,x')  := \exp\bigl( -\g^{-2}\snorm{\P(x)-\P(x')}_H^2  \bigr) 
\end{equation}
defines a new kernel on $X$, see e.g.~\cite[Lemma 4.3]{StCh08}, and 
 if $\P_{\g,Z}:Z\to H_{\g,Z}$ denotes the
canonical feature map
of $k_{\g,Z}$, then $\P_{\g,Z}\circ \P:X\to H_{\g,Z}$ is a feature map of $k_{\g,X, H}$.
Now, note that these assumptions on $\P$ and $H$ are equivalent to saying that 
$H$ is a feature space of the kernel given by $k(x,x') = \langle \P(x), \P(x')\rangle$, and 
since we have $\snorm{\P(x)-\P(x')}_H^2 = k(x,x)-2k(x,x')+k(x',x')$, we can also express 
\eqref{hier-prot-kernel} by the kernel $k$. In the following, we call kernels  $k_{\g,X,H}$ of the form above 
\emph{hierarchical Gaussian kernels}. For later use we note that 
these kernels are universal if $X$ is a compact metric space and $\P$ is continuous and injective,
see Theorem \ref{hier-proto-kernel-universal} for details.

Our next goal is to investigate hierarchical Gaussian kernels \eqref{hier-prot-kernel} for which 
$H$ and its kernel $k$ is 
of rather complicated form. To this end, we write, for 
 $x=(x_1,\dots,x_m)\in \R^m$ and $I\subset \{1,\dots,m\}$,
\begin{displaymath}
  x_I:= (x_i)_{i\in I} 
\end{displaymath}
for the vector projected onto the coordinates listed in $I$.
%
Note that if $X$ is compact then the image 
$X_I$ of this projection is also compact since continuous images of compact sets are compact.
Let us now assume that we have non-empty sets $I_1,\dots,I_l\subset  \{1,\dots,m\}$ 
and $X\subset \R^m$, some weights $w_1,\dots,w_l>0$, 
as well kernels $k_i$ on $X_{I_i}$ for all $i=1,\dots,l$.
For  $I:= I_1\cup \cdots\cup I_l$, we then define a new kernel on $X_I$  by
\begin{equation}\label{weighted-kernel}
   k(x,x') := \sum_{i=1}^l w_i^2 k_i\bigl( x_{I_i}, x'_{I_i} \bigr)\, , \qquad   x,x'\in X_I.
\end{equation}
One can show, that  hierarchical Gaussian kernels of the form \eqref{weighted-kernel} can be computed by a simple product formula, 
see Lemma \ref{hier-kernel-product}. The following definition considers iterations of \eqref{weighted-kernel}.
%


\begin{definition}
 Let $k$ be a kernel of the form \eqref{weighted-kernel} and $H$ be its RKHS. 
 Then the resulting hierarchical Gaussian kernel $k_{\g,X_I,H}$ is said to be of depth
 \begin{enumerate}
  \item  $m= 1$, if all  kernels  $k_1,\dots, k_l$ in  \eqref{weighted-kernel}
  are  linear.
  \item   $m>1$, if all  $k_1,\dots, k_l$ in  \eqref{weighted-kernel}
  are hierarchical Gaussian kernels of depth $m-1$.
 \end{enumerate}
\end{definition}

To illustrate the definition above, we note 
that hierarchical kernels of depth 1 are of the form
\begin{equation}\label{depth-1-kernel}
 k_{\mathbf v}(x,x') :=  \exp\bigl(- \sum_{i\in I} v_i^2 (x_i-x_i')^2\bigr) , \quad   x,x'\in X  ,
\end{equation}
for some suitable $\mathbf v:= (v_i)_{i\in I}$ with 
$v_i> 0$ for all $i\in I$. In other words, they only differ from standard Gaussian kernels on $X_I$ by 
their inhomogeneous width parameter $\mathbf v$. For this reason we sometimes also call 
kernels of the form \eqref{depth-1-kernel} \emph{inhomogeneous Gaussian kernels}.

To derive an explicit formula for depth-2-kernels, we fix some $I_1,\dots,I_l\subset \{1,\dots,d\}$, some
first layer weight vectors
$\mathbf v_1=(v_{i, 1})_{i\in I_1} , \dots,  \mathbf v_l=(v_{i, l})_{i\in I_l} $ and a second layer weight vector 
$\mathbf w=(w_1,\dots w_l)$. Writing $\mathbf W^{(1)}:= (\mathbf v_1,\dots\mathbf v_l)$, 
the  hierarchical Gaussian kernel $k_{\mathbf W^{(1)},  \mathbf w, \g}$  of depth 2 that is build 
upon the kernels $k_{\mathbf v_1}, \dots k_{\mathbf v_l}$ and weights $\mathbf w=(w_1,\dots w_l)$ is given by 
\begin{align}\label{depth-2-kernel}
& k_{\mathbf W^{(1)},  \mathbf w, \g}(x,x') \\\nonumber
 &=  \exp\biggl( -2\g^{-2} \sum_{i=1}^l w_i^2 \bigl(1-  k_{\mathbf v_i}( x_{I_i}, x'_{I_i})\bigr)  \biggr) \\ \nonumber
 &= \exp\Biggl( -2\g^{-2} \sum_{i=1}^l w_i^2 \biggl(1-  \exp\Bigl(- \sum_{j\in I_i} v_{j,i}^2 (x_j-x_j')^2\Bigr) \biggr)  \Biggr)\, .
\end{align}
Here we note that  we used the fact that all first layer kernels  $k_{\mathbf v_1},\dots, k_{\mathbf v_l}$
are normalized, that is $k_{\mathbf v_i}(x,x)=1$ for all $x\in X_{I_i}$. Also note that the parameter $\g$ in 
\eqref{depth-2-kernel} can be consumed by the weights $\mathbf w$. In the following, we therefore use the shorthand 
$k_{\mathbf W^{(1)},  \mathbf w} := k_{\mathbf W^{(1)},  \mathbf w, 1}$.
Finally, if 
$k_{\mathbf W^{(1)}_1,  \mathbf w_1^{(2)}}, \dots, k_{\mathbf W^{(1)}_l,  \mathbf w_l^{(2)}}$ are some hierarchical kernels of depth 2,
and 
$\mathbf w=(w_1,\dots w_l)$ is a third layer weight vector, then the corresponding kernel $k$ of depth 3 is given by 
\begin{align}\label{depth-3-kernel}
 &k_{\mathbf W^{(1)}, \mathbf W^{(2)},\mathbf w, \g}(x,x') \\ \nonumber
 &=  \exp\biggl( -2\g^{-2} \sum_{i=1}^l w_i^2 \bigl(1-  k_{\mathbf W^{(1)}_i,  \mathbf w_i^{(2)}}( x_{I_i}, x'_{I_i})\bigr)  \biggr) \, ,
\end{align}
where we used the notation $\mathbf W^{(1)}:= (\mathbf W^{(1)}_1,\dots\mathbf W^{(1)}_l)$ and $\mathbf W^{(2)}:= (\mathbf w_1^{(2)},\dots\mathbf w_l^{(2)})$.
Repeating these calculations, we see that hierarchical kernels of depth $m$ are given by
\begin{align}\label{depth-m-kernel}
\hspace*{-1.5ex}
 &k_{\mathbf W^{(1)},\dots, \mathbf W^{(m-1)},\mathbf w, \g}(x,x') \\\nonumber
 &=  \exp\biggl( -2\g^{-2} \sum_{i=1}^l w_i^2 
 \bigl(1-  k_{\mathbf W^{(1)}_i,\dots, \mathbf W^{(m-2)}_i,\mathbf w_i^{(m-1)}}( x_{I_i}, x'_{I_i})\bigr)  \biggr) \, ,
\end{align}
where  $k_{\mathbf W^{(1)}_i,\dots, \mathbf W^{(m-2)}_i,\mathbf w_i^{(m-1)}}$ denote 
hierarchical
kernels of depth $m-1$ with $\g=1$. 
Note that because of the recursive definition of hierarchical kernels, 
the number of children kernels $l$ may differ in each parent kernel. 
To be more precise, the kernels $k_{\mathbf W^{(1)}_i,\dots, \mathbf W^{(m-2)}_i,\mathbf w_i^{(m-1)}}$
in \eqref{depth-m-kernel} are  build by $l_i$ many kernels of depth $m-2$, and in general we have both
 $l_i\neq l_j$ and $l_i \neq l$.

\begin{figure}[th]
\centering
\def\layersep{3.5cm}
\begin{tikzpicture}[shorten >=1pt,->,draw=black!50, node distance=\layersep,
scale=0.7, every node/.style={transform shape}]
\tikzstyle{every pin edge}=[<-,shorten <=1pt]

\tikzstyle{node}=[rectangle,draw=black!25,minimum size=17pt,inner sep=3pt]
\tikzstyle{input dim}=[node, circle]
\tikzstyle{final layer}=[node]
\tikzstyle{first layer}=[node]
\tikzstyle{second layer}=[node]
\tikzstyle{col1}=[red]
\tikzstyle{col2}=[blue]
\tikzstyle{col3}=[green]

\newcommand{\myedgeIF}[3]{\draw[col#2] (I-#1.east) -- (F-#2-#3.west) ;}
\newcommand{\myedgeIFt}[3]{\draw[col#2] (I-#1.east) -- (F-#2-#3.west) node[near start,black,anchor=south]{$\big(\mathbf W^{(1)}_{#2}\big)_{#1,#3}$} ;}
\newcommand{\myedgeIFb}[3]{\draw[col#2] (I-#1) -- (F-#2-#3) node[midway,black,anchor=north]{$\big(\mathbf W^{(1)}_{#2}\big)_{#1,#3}$} ;}
\newcommand{\myedgeFEt}[1]{\draw (F-2-#1) -- (E-1) node[near start]{$w_{#1}$} ;}
\newcommand{\myedgeFSt}[2]{\draw (F-#1-#2) -- (S-#1) node[near start]{$w^{(2)}_{#1,#2}$} ;}
\newcommand{\myedgeSEt}[1]{\draw (S-#1) -- (E-1) node[near start]{$w_{#1}$} ;}

\begin{scope}[xshift=3*\layersep]
\foreach \name in {1,...,6} \node[input dim] (I-\name) at (0,5-\name) {$x_{\name}$} ;

\foreach \name in {1,...,4} \node[first layer,col1] (F-1-\name) at (\layersep,5.25-\name / 1.25) {$k_{\big(\mathbf W^{(1)}_1\big)_{\name}}$} ;
\foreach \name in {1,...,4} \node[first layer,col2] (F-2-\name) at (\layersep,2-\name / 1.25) {$k_{\big(\mathbf W^{(1)}_2\big)_{\name}}$} ;

\foreach \name in {1,...,2} \node[second layer] (S-\name) at (2*\layersep,6-3*\name) {$k_{\mathbf W^{(1)}_\name,\mathbf w^{(2)}_\name}$} ;
\node[final layer] (E-1) at (3*\layersep,1.5) {$k_{\mathbf W^{(1)},\mathbf W^{(2)},\mathbf w}$} ;

\myedgeIF123
\myedgeIFt113

\myedgeIF214
\myedgeIF221
\myedgeIF224

\myedgeIF311
\myedgeIF313
\myedgeIF314

\myedgeIF411
\myedgeIF412
\myedgeIF422

\myedgeIF521
\myedgeIF523
\myedgeIF524

\myedgeIF612
\myedgeIFb623

%
%

\foreach \i in {1,...,2} \foreach \j in {1,...,4} \myedgeFSt{\i}{\j} ;

\foreach \i in {1,...,2} \myedgeSEt{\i} ;

\begin{scope}[yshift=5cm]
\draw[-,decorate,decoration={brace,amplitude=3pt},thick] (0.5,0) -- (2.5,0) node[black,midway,anchor=south]{$\mathbf W^{(1)}=(\mathbf W^{(1)}_1,\mathbf W^{(1)}_2)$} ;
\draw[-,decorate,decoration={brace,amplitude=3pt},thick] (4.5,0) -- (6,0) node[black,midway,anchor=south]{$\mathbf W^{(2)}=(\mathbf w^{(2)}_1,\mathbf w^{(2)}_2)$} ;
\draw[-,decorate,decoration={brace,amplitude=3pt},thick] (8,0) -- (9.5,0) node[black,midway,anchor=south]{$\mathbf w$} ;
\end{scope}
\end{scope}

\end{tikzpicture}
\caption{Illustration of a possible hierarchical Gaussian kernel of depth 3.}
\end{figure}
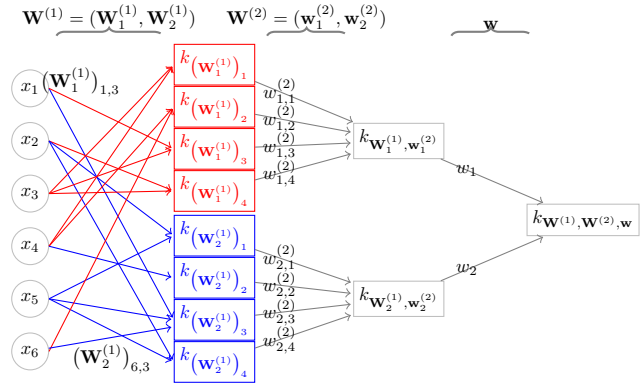

\section{Mathematical Analysis}\label{sec:analysis}

In this section we analyze hierarchical Gaussian kernels theoretically.
Our main result shows that these kernels are universal under a simple
and natural assumption. Based on this, we then show that SVMs using such 
kernels are universally consistent.

\begin{theorem}\label{main-universal-result}
 Let $X\subset \Rd$ be compact. Then 
 every hierarchical Gaussian kernel $k_{\g,X_I,H}$ of some  depth is universal if  $I=\{1,\dots,d\}$.
\end{theorem}

\begin{proofof}{Theorem \ref{main-universal-result}}
Let us first consider the case $I=\{1,\dots,d\}$.
If $k_{\g,X_I,H}$ has depth 1, then the assertion follows by Theorem \ref{hier-kernel-universal} and the 
fact that linear kernels are injective. Moreover, for higher  depths the universality  follows
from Theorem \ref{hier-kernel-universal} and the injectivity of universal kernels. 
\end{proofof}

%
%

Let us fix some $Y\subset \R$.
In the following, a measurable function $L:Y\times \R\to [0,\infty)$ is called a loss.
We further say that the loss $L$ is convex or continuous, if it is convex or continuous with respect to its second argument.
Moreover, $L$ is called Lipschitz continuous, if 
\begin{displaymath}
 |L|_1 := \sup_{y,t\neq t'}\frac{\bigl|L(y,t) -L(y,t')\bigr|}{|t-t'|} < \infty\, ,
\end{displaymath}
where the supremum is taken over all possible values of $y,t$, and $t'$.
Finally, we say that $L$ can be  clipped at some $M>0$ if  
$L(y,\clippt ) \Leq L(y,t)$,
where $\clippt$ denotes the {clipped} value of $t$ at $\pm M$, that is $\clippt := \max\{-M, \min\{M, t\}\}$.
Recall that \cite[Lemma 2.23]{StCh08} gives a simple characterization of convex, clippable losses. In particular, 
the hinge loss is clippable, and so are the 
least squares loss and the pinball loss, if $Y$ is bounded. Furthermore, all these losses are convex and continuous, and 
the hinge loss and the pinball loss are also Lipschitz continuous.
Given a distribution $P$ on $X\times Y$ and a 
(measurable) function $f:X\to \R$, the $L$-risk of $f$ is 
\begin{displaymath}
 \RP L f := \int_{X\times Y}L(y,f(x))\, dP(x,y)\,.
\end{displaymath}
Moreover, the Bayes risk $\RPB L:= \inf\{\RP L f| f:X\to \R\}$ is the smallest possible risk. Finally,  the empirical risk
with respect to some data set $D=((x_1,y_1), \dots,(x_n,y_n))$ is, as usual, 
\begin{displaymath}
 \RD L f := \frac 1 n \sum_{i=1}^nL\bigl(y_i, f(x_i)\bigr)\, .
\end{displaymath}
Now let $L$ be a convex loss, $H$ be an RKHS over $X$ and $\lb>0$ be some regularization parameter. Then the SVM decision function for the data set $D$ is the 
unique solution of the optimization problem 
\begin{equation}\label{svm-opt-problem}
 \fDl = \arg\min_{f\in H} \lb \snorm f_H^2 + \RD L f\, .
\end{equation}
The statistical properties of these learning methods have been extensively studied in the last 15 years, so that a rich theory
is now available, see e.g.~\cite{CuZh07,StCh08}. One key concept of this theory is the so-called  approximation error 
function 
\begin{displaymath}
 A(\lb) := \inf_{f\in H} \lb\snorm f_H^2 + \RP L f -\RPB L\, , \qquad   \lb >0\, ,
\end{displaymath}
which roughly speaking describes the regularization error in an infinite-sample regime. In particular, we have 
$\lim_{\lb\to 0} A(\lb) = 0$ if $X$ is compact, $Y$ is bounded, $H$ is universal, and $L$ is one of the 
losses mentioned above. We refer to \cite[Lemma 5.15 and Corollary 5.29]{StCh08} for the derivation of these results and to 
\cite[Chapter 5.5]{StCh08} for further results in the case of unbounded $Y$.

With the help of these definitions we can now present the following oracle inequality, which is a direct
derivation from  \cite[Theorem 7.22]{StCh08} and the fact that universal kernels have a separable RKHS.

\begin{theorem}\label{consistency-result}
  Let $X\subset \Rd$ be compact, $I=\{1,\dots,d\}$, and   $k_{\g,X_I,H}$ be a 
  hierarchical Gaussian kernel of some depth. Moreover, let $L$ be a Lipschitz continuous loss that can be clipped at some 
  $M>0$ and that satisfies 
  \begin{displaymath}
   L(y,t) \leq B\, , 
  \end{displaymath}
  for some $B>0$ and all $y\in Y$, $t\in [-M,M]$.
  Then there exists a  constant $K>0$
such that for all fixed 
$\t\geq 1$, $n\geq 2$, and $\lb>0$
the SVM associated with $L$ and $k_{\g,X_I,H}$  satisfies 
\begin{align*}
&\RP L \fTc - \RPB L \\
&\quad \leq   9 A(\lb)
+ K   \frac {\ln n}{\lb  n}
+ \frac {15 \t} n \sqrt{\frac{A(\lb)}\lb}
+ \frac{300 B \t} {\sqrt n}
\end{align*}
with probability $P^n$ not less than $1-3e^{-\t}$. In particular, 
the SVM is universally consistent if we pick a regularization sequence $(\lb_n)$
with $\lb_n\to 0$ and $\frac {\ln n}{\lb_n  n}\to 0$, that is
\begin{displaymath}
 \RP L \fTcn \to  \RPB L
\end{displaymath}
in probability for all distributions $P$ on $X\times Y$.
\end{theorem}

Note that the oracle inequality above can also be used to derive learning rates, if, 
as usual, an assumption on the behavior of 
$A(\mycdot)$ is imposed. However, translating such a behavior into 
an assumption on $P$ is even for standard Gaussian kernels
a highly non-trivial task, see e.g.~\cite[Chapter 8.2]{StCh08} and \cite{EbSt13a},
since by \cite{SmZh03a}
all reasonable results in this direction require the kernel parameter $\g$ to change with $n$, too.
While for inhomogeneous kernels there is still some hope to adapt the analysis of \cite{EbSt13a}, 
the situation becomes extraordinarily more difficult for deeper Gaussian kernels. In addition, even 
considering inhomogeneous kernels only,  would most likely be more complicated and lengthy 
than the already involved  analysis of \cite{EbSt13a}, and hence this task is clearly 
out of the scope of this paper. Similarly, our statistical analysis can be sharpened if we add an assumption on the 
entropy number or eigenvalue behavior, see e.g.~\cite[Theorem 7.23 and Chapter 7.5]{StCh08} 
but a refined analysis in this direction is, for essentially the same reasons as above, 
again out of the scope of this paper. Finally, note that some assumptions made in Theorem \ref{consistency-result}
are \emph{not} necessary for deriving consistency. For example, the Lipschitz continuity, the bound in terms of $B$, and the clipping 
assumption can be removed for the price of a looser oracle inequality. For examples in this direction we refer to 
\cite[Chapters 6.4, 7.4, and 9.2]{StCh08}.

\section{Parameter Optimization}\label{sec:param-optimization}

For the standard Gaussian kernels it is well-known both empirically and theoretically
that
the learning performance of the resulting SVM heavily depends on the chosen width parameter $\g$.
Clearly, we can expect a similar dependence on the weight parameters 
${\mathbf W}$ for hierarchical Gaussian kernels, but for these parameters, finding good values 
is a potentially more difficult problem. In this section we address this 
issue by presenting an optimization strategy, which in the next section is empirically validated.

In the following let $D=((x_1,y_1),\dots,(x_n,y_n))$ and $D'=((x'_1,y'_1),\dots,(x'_{n'},y'_{n'}))$ be two data sets. 
Let us assume that $D$ is used for training, that is, $D$ is used to find an SVM decision function
\begin{equation}\label{svm-solution}
 f_{D,\lb,\mathbf W, \g} = \sum_{i=1}^n \a_i k_{\mathbf W, \g}(x_i, \mycdot)\, ,
\end{equation}
where $\mathbf W$ denotes the collection of weights of a hierarchical Gaussian kernel of fixed architecture.
Let us further assume that $D'$ is used to validate the quality of the weights $\mathbf W$, that is,
the empirical risk
\begin{equation}\label{validation-risk}
\Rx L {D'}{f_{D,\lb,\mathbf W, \g}} = \frac 1 {n'} \sum_{j=1}^{n'} L\bigl(y_j',  f_{D,\lb,\mathbf W,\g}(x_j')\bigr)  
\end{equation}
is computed. Clearly, by minimizing this validation risk, we may hope to find a suitable collection of weights $\mathbf W$.
 Note however, that 
 if we change the weights $\mathbf W$ of the kernel $k_{\mathbf W,\g}$ 
 to decrease the validation risk \eqref{validation-risk}, then
 the right-hand side of \eqref{svm-solution} is no longer the SVM solution with respect to the new kernel 
 $k_{\mathbf W,\g}$. Therefore, a recomputation of  $\a:=(\a_1,\dots\a_n)$ becomes necessary, and the same may be true for the 
  hyper-parameters $\lb$ and $\g$. The overall objective is therefore threefold: \emph{a)}
  for fixed $\lb,\g,\a$ we wish to minimize 
 \begin{align}\label{opt-problem}
  &{\ca R}_{L,D'}(\lb, \g, \a, \mathbf W) \\\nonumber
  &\qquad :=   \frac 1 {n'} \sum_{j=1}^{n'} L\biggl(y_j',  \sum_{i=1}^n \a_i  k_{\mathbf W,\g}(x_i, x_j')\biggr) 
 \end{align}
 with respect to $\mathbf W$; \emph{b)} we need to determine $\a$ 
 for fixed $\lb$, $\g$, and $\mathbf W$ by solving the SVM optimization problem \eqref{svm-opt-problem};
 and \emph{c)} we need to update the hyper-parameters $\lb$ and $\g$. 
 Clearly, \emph{b)} is nowadays standard, and for \emph{c)} the most widely adopted approach is grid search based on 
 cross validation. 
 Unfortunately, however, there is no hope that the objective function \eqref{opt-problem} is 
convex in $\mathbf W$, so that minimizing it is   a challenge. In addition, brute-force methods such as  
 grid search over all combinations of weights are far too expensive.  However, finding the `right' weights for learning
 may be a smaller problem compared to, e.g., neural networks, since even for non-optimal weights the corresponding SVM is consistent
 if we split the learning process into two independent phases in which we first look for weights on one (small) chunk of the 
 data and then retrain an SVM with these weights on another (larger) chunk of the data, see 
  Theorem \ref{consistency-result}. In this sense, our task therefore reduces to finding 'good' weights that promise to
  expedite 
 the learning process.

To find such good weights,  we will now present a combination of a local minimizer, namely gradient descent,
 and a global minimization heuristic, namely simulated annealing, but certainly, other approaches such as 
 stochastic gradient descent and its modifications, which are popular for training deep neural networks, may
 be serious alternatives that should be investigated in the future.
 In addition, one could, at least in principle, combine the three objectives \emph{a)}-\emph{c)} into one by a simple summation.
 However, this would change the SVM objective function so that we loose both the well-founded statistical
 theory for SVMs and the ability to use off-the-shelf solvers for \eqref{svm-opt-problem}. For this reason,
 we decided to ignore this possibility. 
 The goal of the remainder of this section is to 
 describe the two optimization  
 strategies for $\mathbf W$ and how they were combined with the steps \emph{b)} and \emph{c)}.

 Let us begin by describing some details of  gradient descent for \eqref{opt-problem}. To this end, let $w$ be a weight 
 contained in $\mathbf W$. If the loss $L$ is differentiable in its second argument, then a simple application of 
 the chain rule shows that 
  the partial derivative of \eqref{opt-problem} with respect to $w$ is given by
 \begin{align*}
 & \frac{\partial}{\partial w}  {\ca R}_{L,D'}(\lb, \g, \a, \mathbf W)\\
  &= \frac 1 {n'} \sum_{j=1}^{n'}  L'\biggl(y_j', \sum_{i=1}^n \a_i  k_{\mathbf W,\g}(x_i, x_j')\biggr)  \sum_{i=1}^n \a_i   \frac{\partial}{\partial w} k_{\mathbf W, \g}(x_i, x_j')  \, ,
 \end{align*}
where $L'$ denotes the derivative of $L$ in its second argument. To implement gradient descent we consequently need to know the 
partial derivatives $ \frac{\partial}{\partial w} k_{\mathbf W,\g}$, which are recursively computed in the following lemma for which the proof and an example can be found in the 
appendix.

\begin{lemma}\label{depth-m-kernel-derivative}
  Let $k_{\mathbf W^{(1)},\dots, \mathbf W^{(m-1)},\mathbf w, \g}$ be a 
  hierarchical Gaussian kernel of depth $m$ 
  of the form \eqref{depth-m-kernel} and $H$ be its RKHS. 
 Then the parameter derivatives of the highest layer can be computed by
 \begin{align*}
 &  \frac{\partial }{\partial w_j}k_{\mathbf W^{(1)},\dots, \mathbf W^{(m-1)},\mathbf w, \g} \\   
 & \qquad  = -4\g^{-2}  w_j  \bigl(1-  k_{\mathbf W^{(1)}_j,\dots, \mathbf W^{(m-2)}_j,\mathbf w_j^{(m-1)}}\bigr) \\  
 &\qquad \quad\cdot   k_{\mathbf W^{(1)},\dots, \mathbf W^{(m-1)},\mathbf w, \g}\, . 
 \end{align*}
 Moreover, the parameter derivative of a weight $w$ occurring in a lower layer of the $j$-th node 
 can be computed  by recursively  using the formula 
 \begin{align*}
  & \frac{\partial }{\partial w}k_{\mathbf W^{(1)},\dots, \mathbf W^{(m-1)},\mathbf w, \g} \\  
  &\qquad = 2\g^{-2} w_j^2
 \frac{\partial }{\partial w}   k_{\mathbf W^{(1)}_j,\dots, \mathbf W^{(m-2)}_j,\mathbf w_j^{(m-1)}}\, . 
 \end{align*}
\end{lemma}

%
%

With the help of the formulas for the partial derivatives the gradient descent step at iteration $i$ now becomes 
\begin{displaymath}
 \mathbf W_{i+1} = \mathbf W_{i} - \eta \nabla \Rx L {D'}{\lb, \g, \a, \mathbf W_i}\, , 
\end{displaymath}
where the gradient $\nabla \Rx L {D'}{\lb, \g, \a, \mathbf W_i}$ is   the vector of partial derivatives and
$\eta$ is the step size. Although the step size is often simply a function of $i$, we 
decided to be more conservative, namely, we determined $\eta$
by a line search based on the Armijo–Goldstein condition, see e.g.~\cite[Chapter 3.5]{BoGiLeSa06}.
This choice ensures that we will find a local minimum, but on the other hand, it certainly
hinders a 
wider inspection of the parameter space. Like for neural networks less conservative approaches
may thus turn out to be more efficient in the future.

%
In this paper we decided to
 address the danger of getting stuck in a poor local minimum of the non-convex objective function \eqref{opt-problem} by
 simulated annealing, see \cite[Chapter 10.12]{PrTeVeFl07}. 
In iteration $i$ of
 our adaptation of this  meta-heuristic 
we randomly picked 
one weight $w$ from the current weights
$\mathbf W_i$, changed it randomly to obtain $\mathbf W_{i+1}$, 
and then kept the change if either 
an improvement of the objective function \eqref{opt-problem} was obtained, that is 
$\D := \Rx L {D'}{\lb, \g, \a, \mathbf W_{i+1}} - \Rx L {D'}{\lb, \g, \a,\mathbf W_i} < 0$, or a
uniformly generated  random number $r_{i+1}\in [0,1]$ satisfied
\begin{displaymath}
 r_{i+1} <
 0.5 \exp\Bigl(- \frac {100 i}{\sqrt N} 
 \cdot \frac{\D}{\Rx L {D'}{\lb, \g, \a,\mathbf W_i}}  \Bigr)\, .
\end{displaymath}
Here $N$ denotes the total number of iterations.

\begin{algorithm}[t]
\caption{Optimization of weights}
\label{opt-algo}
\begin{algorithmic}[1]
{\footnotesize
\REQUIRE A dataset $D$,  a hierarchical Gaussian kernel $k_{\mathbf W}$ of some depth, $L,M, N_1,N_2, N_3\geq 1$.
\ENSURE  Some good values for the weights contained in $\mathbf W$. 
\STATE Split the data set $D$ into three random parts $D_1$, $D_2$, and $D_3$.
\FORALL{$i=1,\dots,M$}
	\STATE Train an SVM $f_{D_1,\lb,\mathbf W}$ of the form \eqref{svm-solution}  
	with $k_{\mathbf W,\g}$ on $D_1$ including  grid search for $\lb$ and $\g$.
          \STATE Optimize $\Rx L {D_2}{\lb, \g, \a, \mathbf W}$ with respect to $\mathbf W$
               by simulated annealing with $N_1$ steps.
	\FORALL{$j=1,\dots,L$}
          \IF {Achieved local minimum of $\Rx L {D_2}{\lb, \g, \a, \mathbf W}$ in previous iteration}
          \STATE Optimize $\Rx L {D_2}{\lb, \g, \a, \mathbf W}$ with respect to $\mathbf W$
               by simulated annealing with $N_2$ steps
          \ELSE
          \STATE Optimize $\Rx L {D_2}{\lb, \g, \a, \mathbf W}$  with respect to $\mathbf W$
              by  gradient descent  with $N_3$ steps.
          \ENDIF
          \STATE Compute the test error $\Rx L {D_3}{\lb, \g, \a, \mathbf W}$ and store $\mathbf W$ if this  error is smaller than the previous one.
	\ENDFOR
	\IF {Test error did not decrease in the loop above}
          \STATE Reshuffle $D_1, D_2$ by splitting $D_1\cup D_2$ into two new parts $D_1$ and $D_2$.
         \ENDIF
\ENDFOR
\STATE \textbf{return} weights $\mathbf W$ that achieved the smallest test error.
}
\end{algorithmic}
\end{algorithm}

In the design of the overall optimization algorithm we again followed a very conservative approach, which 
focuses on reducing  the risk of overfitting. Namely, we split the data set into three parts, from  which the 
first two parts were used to find the SVM decision functions and the objective function \eqref{opt-problem}.
The third part was solely used to track a test error for $\mathbf W$.  The final 
weight $\mathbf W$ was then chosen according to the minimal test error. The corresponding 
algorithm can be found in Algorithm \ref{opt-algo}.

\section{Experiments}\label{sec:experiments}

\begin{table*}[hbtp]
{\small
  \label{raw_error_table}
  \centering  
\begin{tabular}{|r|r|r|r|r|r|r|}
\hline
Data Set & SVM & HKL & Ours & RF & DNN \\
\hline
\!\!{\sc bank}\!\!&\!\!.29780 {\small $\pm$\hspace*{0.2ex}.00240}\!\!&\!\!.29387 {\small $\pm$\hspace*{0.2ex}.00283}\!\!&\!\!\textbf{.25959} {\small $\pm$\hspace*{0.2ex}.00388}\!\!&\!\!\textbf{\textit{.26868}} {\small $\pm$\hspace*{0.2ex}.00269}\!\!&\!\!.29310 {\small $\pm$\hspace*{0.2ex}.00247}\!\! \\
\hline
\!\!{\sc cadata}\!\!&\!\!.05382 {\small $\pm$\hspace*{0.2ex}.00156}\!\!&\!\!.06247 {\small $\pm$\hspace*{0.2ex}.00138}\!\!&\!\!\textbf{\textit{.05252}} {\small $\pm$\hspace*{0.2ex}.00186}\!\!&\!\!\textbf{.05087} {\small $\pm$\hspace*{0.2ex}.00153}\!\!&\!\!.05500 {\small $\pm$\hspace*{0.2ex}.00146}\!\! \\
\hline
\!\!{\sc cod}\!\!&\!\!.15745 {\small $\pm$\hspace*{0.2ex}.00234}\!\!&\!\!.17336 {\small $\pm$\hspace*{0.2ex}.00126}\!\!&\!\!\textbf{\textit{.13093}} {\small $\pm$\hspace*{0.2ex}.00498}\!\!&\!\!.17248 {\small $\pm$\hspace*{0.2ex}.00198}\!\!&\!\!\textbf{.11544} {\small $\pm$\hspace*{0.2ex}.00131}\!\! \\
\hline
\!\!{\sc covtype}\!\!&\!\!.52055 {\small $\pm$\hspace*{0.2ex}.00433}\!\!&\!\!.60997 {\small $\pm$\hspace*{0.2ex}.00418}\!\!&\!\!\textbf{.39955} {\small $\pm$\hspace*{0.2ex}.01484}\!\!&\!\!\textbf{\textit{.48784}} {\small $\pm$\hspace*{0.2ex}.00412}\!\!&\!\!.50274 {\small $\pm$\hspace*{0.2ex}.00629}\!\! \\
\hline
\!\!{\sc cpusmall}\!\!&\!\!.00361 {\small $\pm$\hspace*{0.2ex}.00022}\!\!&\!\!.00460 {\small $\pm$\hspace*{0.2ex}.00041}\!\!&\!\!\textbf{\textit{.00335}} {\small $\pm$\hspace*{0.2ex}.00018}\!\!&\!\!\textbf{.00323} {\small $\pm$\hspace*{0.2ex}.00016}\!\!&\!\!.00375 {\small $\pm$\hspace*{0.2ex}.00015}\!\! \\
\hline
\!\!{\sc cycle}\!\!&\!\!.01048 {\small $\pm$\hspace*{0.2ex}.00035}\!\!&\!\!.01215 {\small $\pm$\hspace*{0.2ex}.00032}\!\!&\!\!\textbf{\textit{.00984}} {\small $\pm$\hspace*{0.2ex}.00047}\!\!&\!\!\textbf{.00838} {\small $\pm$\hspace*{0.2ex}.00031}\!\!&\!\!.01208 {\small $\pm$\hspace*{0.2ex}.00035}\!\! \\
\hline
\!\!{\sc higgs}\!\!&\!\!.90208 {\small $\pm$\hspace*{0.2ex}.00169}\!\!&\!\!.81782 {\small $\pm$\hspace*{0.2ex}.00739}\!\!&\!\!\textbf{\textit{.80235}} {\small $\pm$\hspace*{0.2ex}.01746}\!\!&\!\!\textbf{.77696} {\small $\pm$\hspace*{0.2ex}.00237}\!\!&\!\!.91619 {\small $\pm$\hspace*{0.2ex}.00237}\!\! \\
\hline
\!\!{\sc letter}\!\!&\!\!\textbf{\textit{.04508}} {\small $\pm$\hspace*{0.2ex}.00149}\!\!&\!\!.11505 {\small $\pm$\hspace*{0.2ex}.00176}\!\!&\!\!\textbf{.03394} {\small $\pm$\hspace*{0.2ex}.00136}\!\!&\!\!.05770 {\small $\pm$\hspace*{0.2ex}.00154}\!\!&\!\!\textbf{\textit{.04484}} {\small $\pm$\hspace*{0.2ex}.00185}\!\! \\
\hline
\!\!{\sc magic}\!\!&\!\!.40070 {\small $\pm$\hspace*{0.2ex}.00828}\!\!&\!\!.42824 {\small $\pm$\hspace*{0.2ex}.00819}\!\!&\!\!.38999 {\small $\pm$\hspace*{0.2ex}.00930}\!\!&\!\!\textbf{.37719} {\small $\pm$\hspace*{0.2ex}.00787}\!\!&\!\!\textbf{.37829} {\small $\pm$\hspace*{0.2ex}.00846}\!\! \\
\hline
\!\!{\sc pendigits}\!\!&\!\!\textbf{\textit{.00793}} {\small $\pm$\hspace*{0.2ex}.00075}\!\!&\!\!.02433 {\small $\pm$\hspace*{0.2ex}.00125}\!\!&\!\!\textbf{.00705} {\small $\pm$\hspace*{0.2ex}.00069}\!\!&\!\!.01266 {\small $\pm$\hspace*{0.2ex}.00121}\!\!&\!\!\textbf{\textit{.00789}} {\small $\pm$\hspace*{0.2ex}.00097}\!\! \\
\hline
\!\!{\sc satimage}\!\!&\!\!\textbf{\textit{.04883}} {\small $\pm$\hspace*{0.2ex}.00289}\!\!&\!\!.10783 {\small $\pm$\hspace*{0.2ex}.00588}\!\!&\!\!\textbf{.04670} {\small $\pm$\hspace*{0.2ex}.00300}\!\!&\!\!.05246 {\small $\pm$\hspace*{0.2ex}.00261}\!\!&\!\!.05247 {\small $\pm$\hspace*{0.2ex}.00334}\!\! \\
\hline
\!\!{\sc seismic}\!\!&\!\!.31134 {\small $\pm$\hspace*{0.2ex}.00132}\!\!&\!\!.31893 {\small $\pm$\hspace*{0.2ex}.00221}\!\!&\!\!\textbf{\textit{.29809}} {\small $\pm$\hspace*{0.2ex}.00160}\!\!&\!\!\textbf{.29555} {\small $\pm$\hspace*{0.2ex}.00121}\!\!&\!\!\textbf{\textit{.29746}} {\small $\pm$\hspace*{0.2ex}.00141}\!\! \\
\hline
\!\!{\sc shuttle}\!\!&\!\!.00457 {\small $\pm$\hspace*{0.2ex}.00035}\!\!&\!\!.01293 {\small $\pm$\hspace*{0.2ex}.00071}\!\!&\!\!\textbf{\textit{.00422}} {\small $\pm$\hspace*{0.2ex}.00037}\!\!&\!\!\textbf{.00083} {\small $\pm$\hspace*{0.2ex}.00017}\!\!&\!\!.00593 {\small $\pm$\hspace*{0.2ex}.00038}\!\! \\
\hline
\!\!{\sc thyroid}\!\!&\!\!.17504 {\small $\pm$\hspace*{0.2ex}.00810}\!\!&\!\!.16368 {\small $\pm$\hspace*{0.2ex}.00832}\!\!&\!\!\textbf{\textit{.15380}} {\small $\pm$\hspace*{0.2ex}.00800}\!\!&\!\!\textbf{.02515} {\small $\pm$\hspace*{0.2ex}.00315}\!\!&\!\!\textbf{\textit{.15216}} {\small $\pm$\hspace*{0.2ex}.00803}\!\! \\
\hline
\!\!{\sc updrs}\!\!&\!\!.05372 {\small $\pm$\hspace*{0.2ex}.00517}\!\!&\!\!.17739 {\small $\pm$\hspace*{0.2ex}.00896}\!\!&\!\!\textbf{.00588} {\small $\pm$\hspace*{0.2ex}.00215}\!\!&\!\!\textbf{\textit{.03047}} {\small $\pm$\hspace*{0.2ex}.00160}\!\!&\!\!.05306 {\small $\pm$\hspace*{0.2ex}.00424}\!\! \\
\hline 
\end{tabular}

}
\caption{Average least squares error and standard deviations 
for the considered algorithms on the 15 data sets. For each dataset,   the smallest 
average error is printed in bold face, and the second best error is printed in italics.
In a few cases, the differences were not statistically significant, so that more than two errors are highlighted.}
\end{table*}

While the greater flexibility of hierarchical Gaussian kernels may reduce the 
approximation error, it may also increase the statistical error, that is, the 
danger of overfitting. In addition, 
it is by no means guaranteed, that the optimization strategy described in Section
\ref{sec:param-optimization} is able to find sufficiently good weights.
For this reason we empirically compare our developed method with some other state-of-the-art
learning algorithms.

\textbf{Algorithms.}
As baseline algorithms we picked random forests (RFs) and SVMs since both algorithms 
are well established and 
scored 
best in the recent comparison \cite{FeCeBaAm14a}. In addition, a comparison of our hierarchical
Gaussian kernels to SVMs seems to be somewhat natural.
For the random forests, we simply used the $R$-package \verb randomForest \
and for the  SVMs we found the recent and fast implementation \cite{Steinwart14b}.
Besides these two algorithms, we also   included the
\verb\hierarchical kernel learning (HKL)\
approach of \cite{Bach08a}
and deep neural networks (DNNs) in form of \verb\Caffe\, see \cite{jia2014caffe}.
Since   we scaled the labels of all our data sets to $[-1, 1]$, we 
clipped the final decision functions obtained by the different algorithms to  $[-1, 1]$, too.
Finally, the weight optimization approaches of our algorithm, namely simulated annealing and gradient descent, were
implemented in C++, where the most time-critical inner parts were  ported to GPU code.
Following a conservative approach, all these parts were using double precision although this 
is likely to slow down the GPU code considerably. For the SVM step of our algorithm, we re-used the 
core solver of \cite{Steinwart14b}.

\textbf{Hyper-Parameter-Selection.}
Except RFs the different algorithms come with different hyper-parameters, which need to be 
carefully determined.
For the baseline SVMs we used the default 5-fold cross validation (CV) procedure of \cite{Steinwart14b},
which determines $\lb$ and $\g$ from a 10 by 10 grid. Similarly, we determined the parameter $\lb$
of HKL by 5-fold CV. For the DNNs we choose the architecture from 
15 different candidate architectures with up to 1000 relu-neurons per layer 
by 5-fold CV combined with early stopping, 
and these candidates were 
 determined by some preliminary 
experiments. For our algorithm we considered  inhomogeneous Gaussian 
kernels as well as six hierarchical Gaussian kernels of depth 2 with 4 to 16  nodes in the first layer,
and the final kernel was again chosen by 5-fold CV.
For the weight optimization in Algorithm   \ref{opt-algo}
we split the training sets  
into three parts $D_1$, $D_2$, and $D_3$ 
of size $44.44\%$, $22.22\%$, and    $33.33\%$, respectively. In 
Algorithm \ref{opt-algo} we set  $L=10$, $M=15$, $N_1=1000$, $N_2=500$, and 
$N_3 = 10$. 

All CVs were solely performed by splitting the training data and whenever this was applicable
 the final decision function was obtained by averaging over the resulting five decision functions.
Finally, we ran some preliminary experiments verifying that the setup for the different
algorithms led to satisfying results. In particular for the HKL package we compared 
the results for the data sets {\sc bank-32nh, bank-32nh, pumadyn-32nh, pumadyn-32nh}
reported in \cite{Bach08a} with our setup for the algorithm and obtained 
in all 4 cases slightly better results.

\begin{figure*}[hbtp]
  \label{error_figures}
  \centering  
    \includegraphics[height=0.30\textwidth,width=0.30\textwidth]{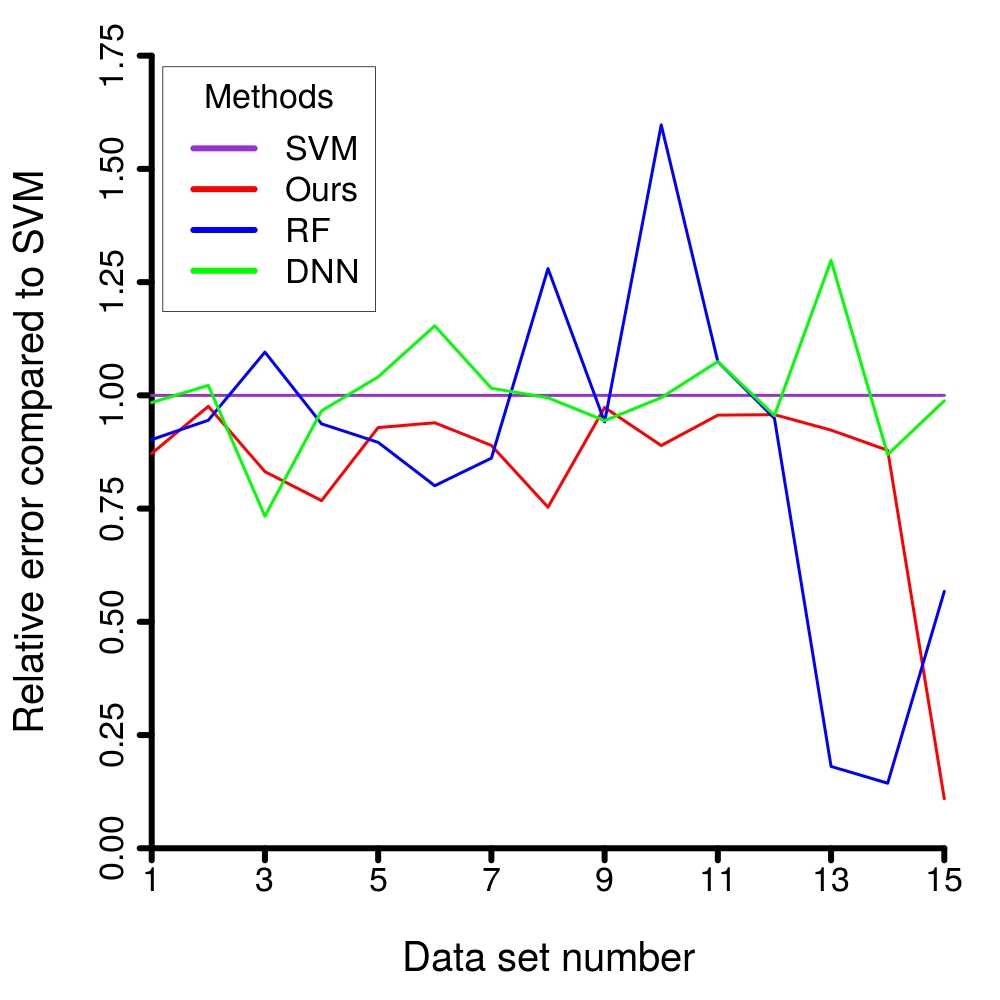}
    \hspace*{0.03\textwidth}
    \includegraphics[height=0.30\textwidth,width=0.30\textwidth]{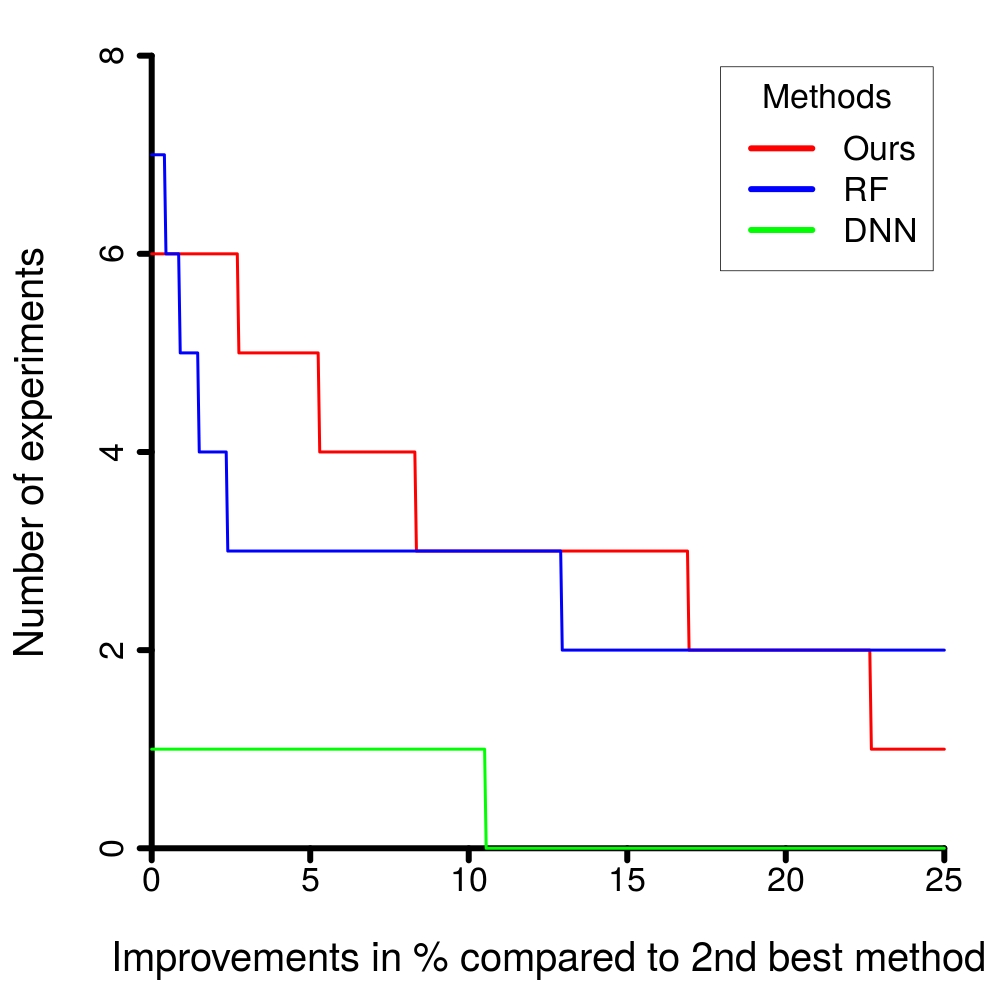}
    \hspace*{0.03\textwidth}
    \includegraphics[height=0.30\textwidth,width=0.30\textwidth]{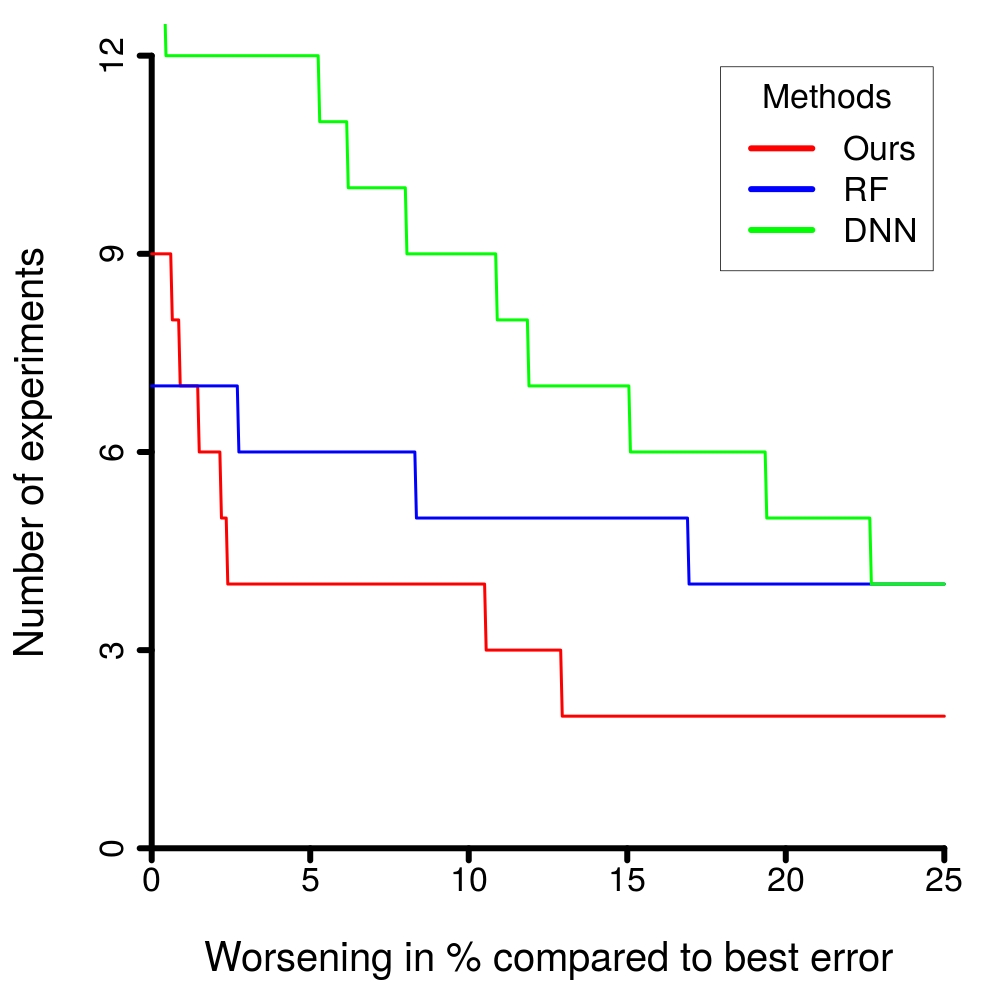}
\caption{Left. Relative errors of the different algorithms on the 15 data sets. 
As a baseline, we picked 
SVMs, that is, the relative error of,  say RFs, on a data set is the ratio
error(RF)/error(SVM). The figure shows that SVMs and DNNs performed equally well,
and that our algorithm consistently outperformed SVMs. The latter is not surprising, since
this was the initial intention of our construction. The figure also shows that 
RFs and DNNs have a quite complementary performance: on most data sets on which 
RFs performed better than the SVM, the DNNs did not, and vice versa.
Middle. This graphic compares our algorithm to RFs and DNNs from a different angle.
It shows, for example, that our algorithm achieved a 5\% better average risk
than the other two algorithms on 5 data sets, while RFs and DNNs did so on
3 and 1 data sets, respectively. This graphic therefore illustrates 
the potential gain of using one of the algorithms.
Right. This graphic illustrates the potential risk of solely using 
one of the three algorithms. For example, it shows that 
our algorithm achieved an error that was 10\% worse than the best error 
on 4 data sets, while RFs and DNNs did so on 5 and 9 data sets, respectively.
}
\end{figure*}

\textbf{Data.}
We downloaded 15 data sets from the UCI Machine Learning Repository and
from LIBSVM's web page, where we note that some  data sets can actually be found on both sites.
In all cases, we scaled the input variables $x$ to $[-1,1]^d$ and the label $y$ to $[-1,1]$.
We randomly generated 30 training and test sets from these data sets, where 
the training set size ranged from 5.000 to 7.000 samples.
The details of the considered data sets as well as their preparation are summarized in Table \ref{data-table}
in the appendix.

\textbf{Execution.}
All experiments were repeated 30 times, so that every algorithm was run 450 times.
These runs were distributed among several computers
with different hardware:
The RF- and SVM-experiments were conducted on an ordinary desktop computer,
where each RF run took between 5 and 60 seconds, and an average SVM run took about 60 seconds.
The DNN runs
were conducted on a PC with an i7-3930K  processor, 64GB of RAM and a Titan GPU. Here, the average experiment
ran for about 45 minutes.
The runs for our algorithm were distributed among 4 identical PCs with 
an i7-5820K processor, 64GB of RAM and a GTX980 GPU. On average, a single run took about 81 minutes, 
but some control experiments on the PC with the Titan GPU showed that on that machine 
an average run would have taken around 45 minutes, too. In addition note, that the DNNs were
using single precision while our algorithm was using double precision, which is between 2 to 3 times 
slower on the Titan. 
Finally, the HKL experiments were run on 10 PCs with i7-3770 processor and 16GB of RAM, and 
an average run took about 60 minutes.
Altogether, we used  more than 1.400 hours, or 58.4 days, of processing time, which indicates that 
the experiments would have been infeasible without distributing the tasks among heterogeneous hardware.

\textbf{Results.}
The average least squares error for each algorithm and data set can be found, 
together with some preliminary ranking of the algorithms, 
in Table 1.
A quick look shows that both our method and RFs 
consistently performed well, scoring on 14, respectively  11, 
data sets among the top two, which are typeset in bold face.
Moreover, our algorithm scored first on 6 data sets, while 
RFs scored best on 7.
The algorithm with the next best scoring is DNN, although this impression
is a bit biased since the SVMs were outperformed by our algorithm, as intended,
on every 
single data set, so that the SVMs had never a chance to score first.
A somewhat better comparison between DNNs and SVMs can thus
be found in the left graphic of Figure 2, which shows that 
both algorithms perform quite comparable. Finally, we were quite
surprised by the performance of HKL, which never scored first or second.
Despite this small issue let us finally have a look at a 
slightly finer comparison between RFs, DNNs and our algorithm, which  can be found in Figure 
2, middle and right. 
Here it turns out that the majority of wins for RFs were
only achieved by a small improvement compared to the second best among these three algorithm,
while our algorithm achieves many of its victories with a larger margin.
For example, RFs won  on 3 data sets by more than 5\% improvement, 
DNNs on 1, and our algorithm even  on 5.
In this respect we finally note, that RFs dramatically 
outperformed all other methods on two data sets, namely {\sc shuttle, thyroid},
while our algorithm did so on {\sc updrs}.
Last but not least, the   graphic on the right hand side of Figure 
2 shows on how many data sets each of the three considered algorithms 
lack behind the best algorithm by which percentage. For example, 
RFs are   5\% worse than the best algorithm on 5   data sets, DNNs on 9,
and our algorithm on 4.

\bibliographystyle{plain}
\bibliography{steinwart-books,steinwart-proc,steinwart-article,steinwart-mine,additional}

\clearpage



\setcounter{section}{0}
\renewcommand{\thesection}{\Alph{section}}


\section{Appendix}\label{sec:appendix}

\begin{lemma}\label{injective-kernel-char}
   For a kernel $k$  on $X$ the following statements are equivalent:
\begin{enumerate}
   \item $k$ is injective.
	\item $k$ has one injective feature map.
\item All feature maps of $k$ are injective.
\end{enumerate}
\end{lemma}

\begin{proofof}{Lemma \ref{injective-kernel-char}}
   The implications $iii) \Rightarrow i) \Rightarrow ii)$ are trivial. To show the remaining implication,
let $\P_0:X\to H_0$ be an arbitrary 
feature map of $k$ and $\P_1:X\to H_1$ be an injective feature map of $k$. For $x,x'$ we then have
\begin{displaymath}
   \snorm{\P_0(x)-\P_0(x')}_{H_0}^2 =   k(x,x)-2k(x,x')+k(x',x')  = \snorm{\P_1(x)-\P_1(x')}_{H_1}^2\, ,
\end{displaymath}
that is $\P_0(x)=\P_0(x')$ if and only if $\P_1(x)=\P_1(x')$, and by assumption the latter is equivalent to 
$x=x'$. 
\end{proofof}

\begin{theorem}\label{hier-proto-kernel-universal}
Let $X$ be a compact metric space and $k$ be a continuous and injective kernel with  feature map $\P:X\to H$ 
and feature space $H$,
then  $k_{\g,X,H}$ defined by \eqref{hier-prot-kernel} is a universal kernel.
\end{theorem}

\begin{proofof}{Theorem \ref{hier-proto-kernel-universal}}
Since $\snorm{\P(x)-\P(x')}_H^2 = k(x,x)-2k(x,x')+k(x',x')$ we may assume without loss of generality that 
$H$ is the RKHS of  $k$ and $\P$ is the canonical feature map of $k$.
Now every compact metric space is separable and since we assumed that $k$ is continuous,
we see by \cite[Lemma 4.33]{StCh08} that $H$ is separable. 
Moreover, the continuity of $k$ implies the continuity of $\P$, see  \cite[Lemma 4.29]{StCh08}, and 
consequently, the assertion follows from 
\cite[Theorem 2.2]{ChSt10a}.
\end{proofof}



\begin{lemma}\label{hier-kernel-product}
 Let $k$ be a kernel of the form \eqref{weighted-kernel} and $H$ be its RKHS. 
 Then the resulting hierarchical Gaussian kernel $k_{\g,X_I,H}$ can be computed by 
 \begin{displaymath}
  k_{\g, X_I,H}(x,x')  = \prod_{i=1}^l k_{\g/w_i, X,H_i}\bigl(x_{I_i},x'_{I_i}\bigr)\, , \qquad\qquad x,x'\in X_I.
 \end{displaymath}
\end{lemma}


\begin{proofof}{Lemma \ref{hier-kernel-product}}
Let $\P:X_I\to H$ be the canonical feature map of $k$, and 
$\P_i:X_{I_i}\to H_i$ be the canonical feature map of $k_i$. Then a    
simple calculation shows that, for all $x,x'\in X_I$, we have 
\begin{align*}
&    
k_{\g, X_I,H}(x,x') \\
 &= \exp\bigl( -\g^{-2}\snorm{\P(x)-\P(x')}_H^2  \bigr) \\
& = \exp\bigl(  -\g^{-2}   k(x,x)+2\g^{-2}k(x,x')-\g^{-2}k(x',x')   \bigr) \\
& = \exp\biggl( -\g^{-2} \sum_{i=1}^l w_i^2 k_i\bigl( x_{I_i}, x_{I_i} \bigr)  
   + 2\g^{-2} \sum_{i=1}^l w_i^2 k_i\bigl( x_{I_i}, x'_{I_i} \bigr) 
   - \g^{-2}\sum_{i=1}^l w_i^2 k_i\bigl( x'_{I_i}, x'_{I_i} \bigr) \biggr) \\
&    = \exp\biggl( -\g^{-2} \sum_{i=1}^l w_i^2\Bigl( k_i\bigl( x_{I_i}, x_{I_i} \bigr) 
- 2  k_i\bigl( x_{I_i}, x'_{I_i} \bigr)  +  k_i\bigl( x'_{I_i}, x'_{I_i} \bigr) \Bigr)\biggr) \\
& =  \exp\biggl( -\g^{-2} \sum_{i=1}^l w_i^2 \mnorm{\P_i(x_{I_i})-\P_i(x'_{I_i})}_{H_i}^2 \biggr) \\
& =  \prod_{i=1}^l \exp\biggl( -\g^{-2}  w_i^2 \mnorm{\P_i(x_{I_i})-\P_i(x'_{I_i})}_{H_i}^2 \biggr) \\
& = \prod_{i=1}^l k_{\g/w_i, X,H_i}\bigl(x_{I_i},x'_{I_i}\bigr)\, .
\end{align*}
In other words, we have shown the assertion.
\end{proofof}


\begin{theorem}\label{hier-kernel-universal}
 Let $k$ be a kernel of the form \eqref{weighted-kernel} and $H$ be its RKHS. 
If all kernels $k_1,\dots,k_l$ are injective, and $X_I$ is compact, then 
the resulting hierarchical Gaussian kernel $k_{\g,X_I,H}$  is universal.
\end{theorem}


\begin{proofof}{Theorem \ref{hier-kernel-universal}}
By Theorem \ref{hier-proto-kernel-universal} and the assumed injectivity of $k_i$
we know  that  $k_{\g/w_i, X_i,H_i}$ is universal on $X_{I_i}$. 
Moreover, Lemma \ref{hier-kernel-product} shows that 
 \begin{displaymath}
  k_{\g, X,H}(x,x')  = \prod_{i=1}^l k_{\g/w_i, X,H_i}\bigl(x_{I_i},x'_{I_i}\bigr)\, , \qquad\qquad x,x'\in X.
 \end{displaymath}
Therefore, Lemma \ref{tensor-universal}
together with a simple induction over $l$  gives the desired result.
%
%
%
%
\end{proofof}

\begin{lemma}\label{tensor-universal}
Let $X\subset \R^m$ be a compact and non-empty subset, $I,J\subset \{1,\dots,m\}$ be non-empty,
and $k_I$ and $k_J$ be universal kernels on $X_I$ and $X_J$, respectively. Then 
$k_I\otimes k_J$ defined by 
\begin{displaymath}
k_I\otimes k_J(x , x' ) := k_I(x_I, x'_I) \cdot k_J(x_J, x'_J)   
\end{displaymath}
for all  $x , x' \in X_{I\cup J}$
  is a universal kernel on $X_{I\cup J}$.
\end{lemma}

\begin{proofof}{Lemma \ref{tensor-universal}}
Let us denote the RKHSs of $k_I$ and $k_J$ by $H_I$ and $H_J$, respectively. 
Similarly, we write $\P_I$ and $\P_J$ for the canonical feature maps of these kernels.
Moreover, let $H_I \otimes H_J$ be the formal tensor product of $H_I$ and $H_J$. 
Recall that this tensor product is a vector space spanned by the elementary tensors
$h_I\otimes h_J$, where $h_I\in H_I$ and $h_J\in H_J$. Moreover, there is a unique 
inner product $\langle \mycdot,\mycdot\rangle_{H_I \otimes H_J}$ on $H_I \otimes H_J$ that satisfies 
\begin{displaymath}
   \langle h_I\otimes h_J, h'_I\otimes h'_J \rangle_{H_I \otimes H_J} = \langle h_I, h'_I\rangle_{H_I} \cdot \langle h_J, h'_J\rangle_{H_J} 
\end{displaymath}
for all elementary tensors. We write $H_I\hat\otimes H_J$ for the completion of $H_I \otimes H_J$
with respect to the norm resulting from $\langle \mycdot,\mycdot\rangle_{H_I \otimes H_J}$.
Then the map 
\begin{align*}
 \P_I\otimes \P_J:X_{I\cup J}& \to H_I\hat\otimes H_J \\  
x&\mapsto \P_I(x_I) \otimes \P_J(x_J)
\end{align*}
 is a feature map of $k_I\otimes k_J$
since for $x,x'\in  X_{I\cup J}$ we have  
\begin{align*}
   \bigl\langle \P_I\otimes \P_J(x) , \P_I\otimes \P_J(x')\bigr\rangle_{H_I\hat\otimes H_J}
& =  \bigl\langle \P_I(x_I)\otimes \P_J(x_J) , \P_I(x'_I)\otimes \P_J(x'_J)\bigr\rangle_{H_I\otimes H_J}\\
& =  \bigl\langle \P_I(x_I), \P_I(x'_I)\bigr\rangle_{H_I} 
			\cdot  \bigl\langle \P_J(x_J) , \P_J(x'_J)\bigr\rangle_{H_J} \\
& = k_I(x_I, x'_I) \cdot k_J(x_J, x'_J)  \\
& = k_I\otimes k_J(x , x' ) \, .
\end{align*}
By \cite[Theorem 4.21]{StCh08} we then know that the RKHS $H$ of $k_I\otimes k_J$ is given by 
\begin{displaymath}\label{almost-universal}
   H = \Bigl\{ h:X_{I\cup J}\to \R  \,\bigl|\,  \exists v\in   H_I\hat\otimes H_J \mbox{ with } h(x) = \bigl\langle v, \P_I\otimes \P_J(x)\bigr\rangle_{H_I\hat\otimes H_J} \mbox{ for all $x\in X_{I\cup J}$}   \Bigr\}\, .
\end{displaymath}
Now observe that $k_I$ and $k_J$ are continouos because of  their assumed universality  
and therefore, $k_I\otimes k_J$ is continuous, too. In particular, $H$ consists of continuous functions.

Let us now consider the space
\begin{displaymath}
   C(X_I) \otimes C(X_J) := \spann \bigl\{ f\otimes g\, \bigl| \, f\in C(X_I),\, g\in C(X_J)  \bigr\}\, ,
\end{displaymath}
where again we define the functions $f\otimes g:X_{I\cup J}\to \R$ 
by $f\otimes g(x) := f(x_I)\cdot g(x_J)$ for all $x\in X_{I\cup J}$. 
Our next goal is to show that $H$ approximates the space  $C(X_I) \otimes C(X_J)$ arbitrarily well
with respect to the $\inorm\cdot$, that is
\begin{equation}
  C(X_I) \otimes C(X_J) \subset \overline{H}^{\inorm\cdot}\, ,
\end{equation}
where $\overline{H}^{\inorm\cdot}$ denotes the $\inorm\cdot$-closure of $H$ in $C(X_{I\cup J})$.
To this end, it clearly suffices to show that each $f\otimes g\in  C(X_I) \otimes C(X_J)$ 
can be arbitrarily well approximated. To show the latter, we fix an $\e>0$. Since $k_I$ and $k_J$ are universal,
there then exist $h_I\in H_I$ and $h_J\in H_J$ with $\inorm {h_I-f}\leq \e$ and $\inorm {h_J-g}\leq \e$.
For the function $h\in H$ defined by
$h(x):= \bigl\langle h_I\otimes h_J, \P_I\otimes \P_J(x)\bigr\rangle_{H_I\hat\otimes H_J}$ we then have 
\begin{align*}
   h(x) = \bigl\langle h_I\otimes h_J, \P_I\otimes \P_J(x)\bigr\rangle_{H_I\hat\otimes H_J} 
= \bigl\langle h_I, \P_I(x_I)\bigr\rangle_{H_I}
\cdot \bigl\langle h_J, \P_J(x_J)\bigr\rangle_{H_J}
= h_I(x_I) \cdot h_J(x_J)
\end{align*}
for all $x\in X_{I\cup J}$. This equation together with $\inorm{h_J} - \inorm{g} \leq\inorm{h_J-g}\leq \e $ now gives
\begin{align*}
   \inorm {h-f\otimes g} = \inorm{h_Ih_J - fg}
&\leq \inorm{h_Ih_J - fh_J}   + \inorm{fh_J - fg} \\
&\leq \inorm{h_I-f}\cdot \inorm{h_J } + \inorm{f}\cdot\inorm{h_J-g} \\
& \leq \e \cdot (\inorm g+\e) + \inorm f \cdot\e \, .
\end{align*}
From the latter we easily conclude that \eqref{almost-universal} holds.

In view of \eqref{almost-universal} we now show in the last step of the 
proof that $\ca A:=C(X_I) \otimes C(X_J)$ is dense in $C(X_{I\cup J})$.
To this end, we first observe that $\ca A$ is a sub-algebra of $C(X_{I\cup J})$ by construction. 
Moreover, $\ca A$ does not vanish, since for $x\in X_{I\cup J}$ there exist $f\in C(X_I)$ and $g\in C(X_J)$
with $f(x_I) \neq 0$ and $g(x_J) \neq 0$. This gives $f\otimes g(x) = f(x_I)\cdot g(x_J)\neq 0$.
Finally, $\ca A$ also separates points.  To check this, we first observe that 
for $x,x'\in X_{I\cup J}$ with $x\neq x'$ 
we have $x_I\neq x_I'$ or $x_J\neq x'_J$. Let us assume without loss of generality that $x_I\neq x_I'$.
Then there exists an $f\in C(X_I)$ with $f(x_I)\neq f(x'_J)$. For $g:=\eins_{X_J}\in C(X_J)$ 
being the constant one function, we then 
obtain $f\otimes g(x) = f(x_I) \neq f(x'_I) = f\otimes g(x')$,
and therefore $\ca A$ does separate points. By the theorem of Stone-Weierstra\ss\ we then see that $\ca A$ is dense 
in $C(X_{I\cup J})$.
\end{proofof}

\begin{lemma}\label{kernel-derivative}
  Let $k$ be a kernel of the form \eqref{weighted-kernel} and $H$ be its RKHS. 
 Then the resulting hierarchical Gaussian kernel $k_{\g,X_I,H}$ satisfies
 \begin{displaymath}
   \frac{\partial }{\partial w_j}k_{\g, X_I,H}(x,x')  = -2w_j \g^{-2} \mnorm{\P_j(x_{I_j})-\P_j(x'_{I_j})}_{H_j}^2 k_{\g, X_I,H}(x,x') \, , \qquad\qquad x,x'\in X.
 \end{displaymath}
\end{lemma}

\begin{proofof}{Lemma \ref{kernel-derivative}}
Using Lemma \ref{hier-kernel-product} we find
 \begin{align*}
   \frac{\partial }{\partial w_j} k_{\g, X_I,H}(x,x') 
   & =  \frac{\partial }{\partial w_j} \prod_{i=1}^l k_{\g/w_i, X,H_i}\bigl(x_{I_i},x'_{I_i}\bigr) \\
  & =    \prod_{i\neq j} k_{\g/w_i, X,H_i}\bigl(x_{I_i},x'_{I_i}\bigr)  
      \cdot   \frac{\partial }{\partial w_j} k_{\g/w_j, X,H_j}\bigl(x_{I_j},x'_{I_j}\bigr)  \, .
 \end{align*}
 Moreover, we have 
 \begin{align*}
  & \frac{\partial }{\partial w_j} k_{\g/w_j, X,H_j}\bigl(x_{I_j},x'_{I_j}\bigr)  \\
   &= \frac{\partial }{\partial w_j} \exp\biggl( -\g^{-2}  w_j^2 \mnorm{\P_j(x_{I_j})-\P_j(x'_{I_j})}_{H_j}^2 \biggr) \\
   & = - 2\g^{-2}  w_j \mnorm{\P_j(x_{I_j})-\P_j(x'_{I_j})}_{H_j}^2 \exp\biggl( -\g^{-2}  w_j^2 \mnorm{\P_j(x_{I_j})-\P_j(x'_{I_j})}_{H_j}^2 \biggr) \\
   & = - 2\g^{-2}  w_j \mnorm{\P_j(x_{I_j})-\P_j(x'_{I_j})}_{H_j}^2  k_{\g/w_j, X,H_j}\bigl(x_{I_j},x'_{I_j}\bigr) \, ,
 \end{align*}
and by combining both expressions we obtain the assertion.
\end{proofof}

\begin{proofof}{Lemma \ref{depth-m-kernel-derivative}}
Let $H_j$ be a RKHS of $k_{\mathbf W^{(1)}_j,\dots, \mathbf W^{(m-2)}_j,\mathbf w_j^{(m-1)}}$
and $\P_j$ be the corresponding canonical feature map.
 Then the  formula for the highest layer follows from Lemma \ref{kernel-derivative} and 
 \begin{displaymath}
  \mnorm{\P_j(x_{I_j})-\P_j(x'_{I_j})}_{H_j}^2 
  = 2 - 2 k_{\mathbf W^{(1)}_j,\dots, \mathbf W^{(m-2)}_j,\mathbf w_j^{(m-1)}}(x_{I_j},x_{I_j}')\, .
 \end{displaymath}
Moreover, if $w$ is a weight that occurs in a lower layer of the $j$-th node, that is, it occurs at 
the component $\mathbf W^{(n)}_j$ of $\mathbf W^{(n)}$ for some $n\leq m-1$, then 
\eqref{depth-m-kernel} gives
\begin{align*}
  \frac{\partial }{\partial w} k_{\mathbf W^{(1)},\dots, \mathbf W^{(m-1)},\mathbf w, \g}(x,x') 
  & = \frac{\partial }{\partial w} \exp\biggl(\!\! -2\g^{-2} \sum_{i=1}^l w_i^2 
 \bigl(1\!-\!  k_{\mathbf W^{(1)}_i,\dots, \mathbf W^{(m-2)}_i,\mathbf w_i^{(m-1)}}( x_{I_i}, x'_{I_i})\bigr) \!\! \biggr) \\
 &=  \exp\biggl( -2\g^{-2} \sum_{i\neq j} w_i^2 
 \bigl(1-  k_{\mathbf W^{(1)}_i,\dots, \mathbf W^{(m-2)}_i,\mathbf w_i^{(m-1)}}( x_{I_i}, x'_{I_i})\bigr)  \biggr) \\
& \quad \cdot \frac{\partial }{\partial w} \exp\biggl(\! -2\g^{-2} w_j^2 
 \bigl(1-  k_{\mathbf W^{(1)}_j,\dots, \mathbf W^{(m-2)}_j,\mathbf w_j^{(m-1)}}( x_{I_j}, x'_{I_j})\bigr) \! \biggr)\, .  
\end{align*}
In addition, the chain rule yields
\begin{align*}
 &\frac{\partial }{\partial w} \exp\biggl(\! -2\g^{-2} w_j^2 
 \bigl(1-  k_{\mathbf W^{(1)}_j,\dots, \mathbf W^{(m-2)}_j,\mathbf w_j^{(m-1)}}( x_{I_j}, x'_{I_j})\bigr) \! \biggr)\\
 & = \exp\biggl(\! -2\g^{-2} w_j^2 
 \bigl(1-  k_{\mathbf W^{(1)}_j,\dots, \mathbf W^{(m-2)}_j,\mathbf w_j^{(m-1)}}( x_{I_j}, x'_{I_j})\bigr) \! \biggr) \\
&\quad \cdot
 \frac{\partial }{\partial w}  
 \Bigl(-2\g^{-2} w_j^2 \bigl(1-  k_{\mathbf W^{(1)}_j,\dots, \mathbf W^{(m-2)}_j,\mathbf w_j^{(m-1)}}( x_{I_j}, x'_{I_j})\bigr) \Bigr)\\
 & = \exp\biggl(\! -2\g^{-2} w_j^2 
 \bigl(1-  k_{\mathbf W^{(1)}_j,\dots, \mathbf W^{(m-2)}_j,\mathbf w_j^{(m-1)}}( x_{I_j}, x'_{I_j})\bigr) \! \biggr) \\
&\quad \cdot 2 \g^{-2} w_j^2
 \frac{\partial }{\partial w}   k_{\mathbf W^{(1)}_j,\dots, \mathbf W^{(m-2)}_j,\mathbf w_j^{(m-1)}}( x_{I_j}, x'_{I_j})\, .
\end{align*}
Combining both expressions, we then obtain the second formula.
\end{proofof}

To illustrate the lemma~\ref{depth-m-kernel-derivative} we note that the parameter 
derivatives of hierarchical Gaussians $k_{\mathbf v}$ of depth 1, that is, of kernels of 
the form \eqref{depth-1-kernel}, are given by 
\begin{displaymath}
 \frac{\partial }{\partial v_j} k_{\mathbf v} (x,x') 
 =  -2v_j (x_j-x_j')^2  k_{\mathbf v} (x,x') \, , \qquad \qquad x,x'\in X\, .
\end{displaymath}
To derive an explicit formula for depth-2-kernels we fix 
some
first layer weight vectors
$\mathbf v_1=(v_{i, 1})_{i\in I_1} , \dots,  \mathbf v_l=(v_{i, l})_{i\in I_l} $ and a second layer weight vector 
$\mathbf w=(w_1,\dots w_l)$. Let us write  $\mathbf V:=\mathbf W^{(1)} := (\mathbf v_1,\dots\mathbf v_l)$.
The first formula of Lemma \ref{depth-m-kernel-derivative} then shows that 
the depth-2 kernel 
$k_{\mathbf V,  \mathbf w, \g}$ defined in \eqref{depth-2-kernel} has the $\mathbf w$-parameter derivatives
\begin{align*}
 \frac{\partial }{\partial w_j} k_{\mathbf V,  \mathbf w, \g}(x,x')  
 = -4w_j \g^{-2} \bigl( 1 - k_{\mathbf v_j} (x,x')\bigr) k_{\mathbf V,  \mathbf w, \g}(x,x')  \, .
\end{align*}
Moreover, the second formula of Lemma \ref{depth-m-kernel-derivative} gives
\begin{align*}
 \frac{\partial}{\partial v_{i,j}} k_{\mathbf V,  \mathbf w, \g}(x,x')  
 = 2 \g^{-2} w_j^2
 \frac{\partial }{\partial v_{i,j}}   k_{\mathbf v_j}(x,x') 
 = - 4 v_{i,j}\g^{-2} w_j^2  (x_j-x_j')^2  k_{\mathbf v_j} (x,x')\, .
\end{align*}



\begin{table}
\begin{center}
{
{\footnotesize
\begin{tabular}{|r|r|r|r|r|r|r|}
\hline
Data Set &  Size & Dimension. & Type & Training & Testing  &Source\\
\hline
{\sc Bank-Marketing}  &  45211  &  16   &  BC & 7000 & 24212 & UCI \\
\hline
{\sc Cadata}  & 20640   &   8  & REG &  7000 & 6640 & LIBSVM \\
\hline
{\sc Cod-rna}  & 331152   &  8   &  BC & 7000 & 310154 & LIBSVM \\
\hline
{\sc Covtype}  & 581012   & 55    &  BC & 7000& 560014 & UCI \\
\hline
{\sc Cpusmall}  &  8192  &  12   & REG &  6000& 2192 & LIBSVM \\
\hline
{\sc Cycle-Power-Plant}  &  9568  &  4   & REG & 6000 & 3568  & UCI \\
\hline
{\sc Higgs}  &  500000  &  28   &   BC &7000 & 479002 & UCI \\
\hline
{\sc Letter-Recognition}  &  20000  &  16   & MC-26 & 7000& 6000&  UCI   \\
\hline
{\sc Magic}  & 19020   &   10  & BC & 7000 & 5022 & UCI   \\
\hline
{\sc Pendigits}  & 10992   &  16   & MC-10 & 7000 & 3992 & UCI  \\
\hline
{\sc Satimage}  & 6435   &   36  &  MC-6  & 5000 & 1435 & UCI\\
\hline
{\sc Seismic}  &  98528  & 50  & MC-3  & 7000 & 77530 & LIBSVM   \\
\hline
{\sc Shuttle}  &  43500  & 8    & MC-6 & 7000 & 22500 & LIBSVM    \\
\hline
{\sc Thyroid-ann}  & 7200   &  21   &  BC & 5000 & 2001 & UCI \\
\hline
{\sc Updrs-motor}  & 5875   &  19   &  REG & 5000 & 875 & UCI \\
\hline
\end{tabular}
}
}
\end{center}
\caption{{\small Details of the considered datasets and their preparation. All data sets were scaled 
to  $[-1,1]^{d+1}$ or $[0,1]$. For {\sc Shuttle} we additionally removed the first 
data column that contains a time stamp whose influence has been debated in the literature.
The types are binary classification (BC), regression (REG), and multiclass classification with $k$ labels
(MC-$k$). We subsampled training and test sets of the specified size. For the larger datasets we initially
planned some additional experiments, for which a separate data set would have been needed.
This explains the difference between the total size and the sum of the training and test set sizes.
These splits  were 30 times repeated.
}}
\label{data-table}
\end{table}

\end{document}